%% file: main.tex
\documentclass[letterpaper]{article} 
\usepackage{aaai25}  
\usepackage{times}  
\usepackage{helvet}  
\usepackage{courier}  
\usepackage[hyphens]{url}  

\usepackage[table]{xcolor}
\definecolor{DeepGreen}{RGB}{0, 200, 0}
\definecolor{mydarkred}{rgb}{0.6,0,0}
\definecolor{myblue}{HTML}{268BD2}
\definecolor{mygreen}{HTML}{658354}

\usepackage{pgfplots}
\usepackage{graphicx}
\usepackage{tikz}
\usetikzlibrary{trees}

\usepackage{float}
\usepackage[labelformat=simple]{subfig}
\usepackage{booktabs}
\usepackage{array}
\usepackage{multirow}
\usepackage{makecell}
\usepackage{tabularx}
\usepackage{colortbl}
\usepackage{siunitx}
\usepackage{varwidth}
\usepackage{hhline}

\usepackage{amsfonts}
\usepackage{amssymb}
\usepackage{amsmath}
\usepackage{bm}
\usepackage{nicefrac}
\usepackage{microtype}

\usepackage{cleveref}
\usepackage{pdfpages}
\usepackage{algorithm}
\usepackage{algorithmic}
\usepackage{listings}
\usepackage{newfloat}
\usepackage{pifont}
\usepackage{nicematrix}
\usepackage{natbib}
\usepackage{caption}
\usepackage{natbib}  
\usepackage{caption} 
\usepackage{algorithm}
\usepackage{algorithmic}
\usepackage{makecell}
\usepackage{xcolor}
\usepackage{tikz}
\usetikzlibrary{trees}
\definecolor{DeepGreen}{RGB}{0, 200, 0}
\usepackage{newfloat}
\usepackage{listings}
\usepackage{amssymb}
\usepackage{pifont}
\newcommand{\cmark}{\ding{51}}%

\definecolor{mydarkred}{rgb}{0.6,0,0}
\definecolor{myblue}{HTML}{268BD2}
\definecolor{mygreen}{HTML}{658354}

\DeclareCaptionStyle{ruled}{labelfont=normalfont,labelsep=colon,strut=off} 
\floatstyle{ruled}
\newfloat{listing}{tb}{lst}{}
\floatname{listing}{Listing}
\title{Out-of-Distribution Detection with Prototypical Outlier Proxy}
\author{
    Mingrong Gong\textsuperscript{\rm 1},
    Chaoqi Chen\textsuperscript{\rm 1}\thanks{Corresponding author.},
    Qingqiang Sun\textsuperscript{\rm 2},
    Yue Wang\textsuperscript{\rm 3},
    Hui Huang\textsuperscript{\rm 1}
}
\affiliations{
    \textsuperscript{\rm 1}Shenzhen University,\;
    \textsuperscript{\rm 2}Great Bay University,\;
    \textsuperscript{\rm 3}University College London\\

\{gmr52333, cqchen1994, hhzhiyan\}@gmail.com,\; qqsun@gbu.edu.cn,\; zcabwaa@ucl.ac.uk
}

\usepackage{bibentry}

\begin{document}

\maketitle

\input{sections/1.abstract.tex}
\input{sections/2.introduction.tex}

\input{sections/3.motivation}
\input{sections/4.methods}
\input{sections/5.experiments}

\input{sections/6.related_work}
\input{sections/7.discussion.tex}
\bigskip
\bibliography{aaai25}
\input{sections/appendix}
\end{document}

%% file: sections/1.abstract.tex
\begin{abstract}
Out-of-distribution (OOD) detection is a crucial task for deploying deep learning models in the wild. 
One of the major challenges is that well-trained deep models tend to perform over-confidence on unseen test data. 
Recent research attempts to leverage real or synthetic outliers to mitigate the issue, which may significantly increase computational costs and be biased toward specific outlier characteristics.
In this paper, we propose a simple yet effective framework, \textit{Prototypical Outlier Proxy} (POP), which introduces virtual OOD prototypes to reshape the decision boundaries between ID and OOD data.
Specifically, we transform the learnable classifier into a fixed one and augment it with a set of prototypical weight vectors.
Then, we introduce a hierarchical similarity boundary loss to impose adaptive penalties depending on the degree of misclassification. 
Extensive experiments across various benchmarks demonstrate the effectiveness of POP. Notably, POP achieves average FPR95 reductions of 7.70\%, 6.30\%, and 5.42\% over the second-best methods on CIFAR-10, CIFAR-100, and ImageNet-200, respectively. Moreover, compared to the recent method NPOS, which relies on outlier synthesis, POP trains 7.2$\times$ faster and performs inference 19.5$\times$ faster. 
The source code is available at: https://github.com/gmr523/pop.
\end{abstract}

%% file: sections/2.introduction.tex
\section{Introduction}
\label{sec:intro}
\begin{figure}
    \centering
    \includegraphics[width=1\linewidth]{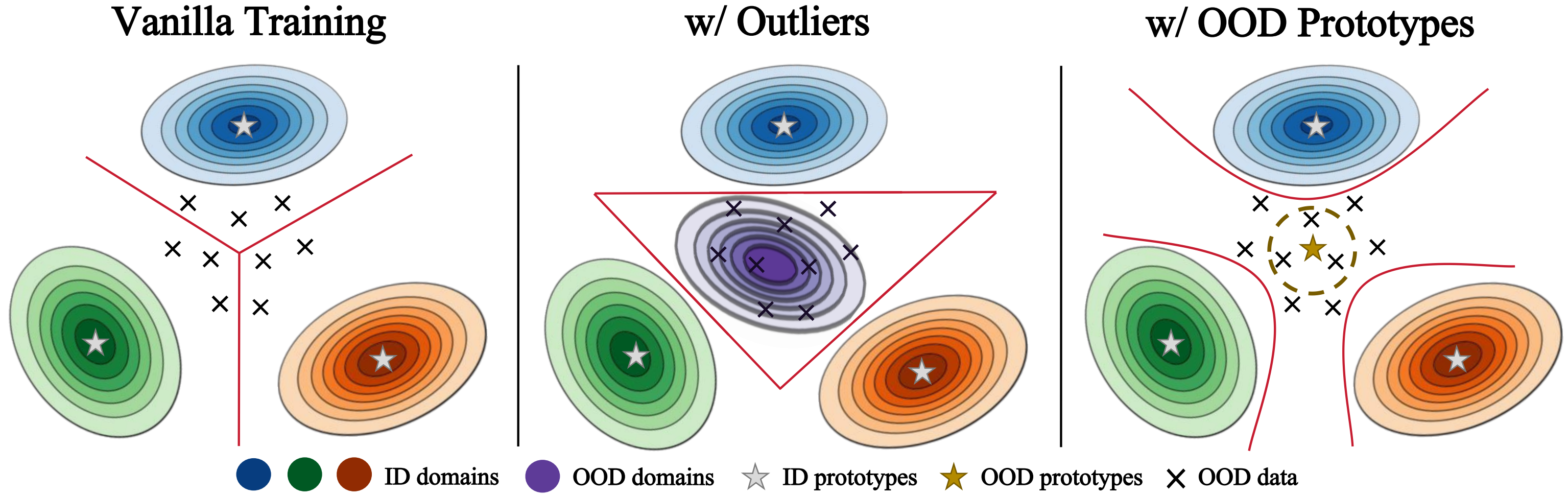}
    \caption{Illustration of our motivation. 
    \textit{Left}: Vanilla training. \textit{Middle}: Training with the mixture of ID data and outliers. \textit{Right}: Training with prototypical outlier proxies.}
    \label{fig:teaser}
\vspace{-5mm}
\end{figure}

Deep learning models have achieved remarkable success across various tasks such as image classification~\cite{He_2016_CVPR}, face recognition ~\cite{deng2019arcface}, and object detection ~\cite{object}. 
However, the safety requirements of these models pose significant challenges when deployed in real-world applications, such as autonomous driving~\cite{drive}, robotics~\cite{robot}, and medical diagnostics~\cite{amodei2016concrete}. 
Albeit the extraordinary performance on in-distribution (ID) data, such models struggle to deal with out-of-distribution (OOD) data,
which may result in misclassifications, misguided decisions, and even catastrophe.
As shown in Fig. \ref{fig:teaser} (left), to achieve higher training accuracy, deep models tend to make overconfident predictions~\cite{DBLP:conf/icml/GuoPSW17}, even in the low-density regions.
To solve this issue, many existing methods strive to directly introduce outliers to enhance the unknown-aware ability during the training phase, using either real outlier data, \textit{i.e.,} outlier exposure~(OE)~\cite{DBLP:conf/iclr/HendrycksMD19, Yu2019UnsupervisedOD, DBLP:conf/iccv/YangWFYZZ021,ming2022poem,DBLP:conf/wacv/ZhangILCL23}, or feature-based outlier synthesis~\cite{Pei2021OutofdistributionDW, DBLP:conf/iclr/DuWCL22VOS, DBLP:conf/iclr/TaoDZ023}. 
Fig. \ref{fig:teaser} (middle) shows that training with outliers will create a specific region to accommodate potential OOD data.
Despite the promise, these methods may still be constrained by two bottlenecks: 
(\romannumeral1) Incorporating extra outliers in the training phase can be time-consuming and resource-intensive. 
For example, synthesizing outliers requires density estimation (parametric~\cite{DBLP:conf/iclr/DuWCL22VOS} or non-parametric~\cite{DBLP:conf/iclr/TaoDZ023}) of ID data first.
(\romannumeral2) In practice, OOD data are diverse and typically distribution-free~\cite{fang2022out}. 
Thus, OE methods may only be effective in certain specific domains because it is impossible to cover all potential scenarios. These methods may cause the model to be biased towards specific outlier characteristics, leading to a loss of generality. For instance, the model might learn spurious correlations between the data and binary labels~\cite{ming2022impact}.
To this premise, we raise an open question: 
\begin{center}
    \textit{Can we enable deep models to perceive unseen data without introducing any specific outliers?}
\end{center}

\begin{figure}[t]
    \centering
    \includegraphics[width=0.85\linewidth]{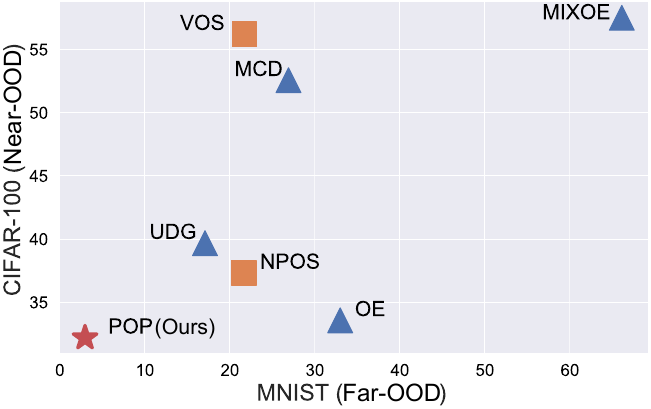}
    \caption{FPR95 (\%) of six OOD detection baselines and our POP, using ResNet-18 trained on CIFAR-10, tested on CIFAR-100 and MNIST. Lower FPR95 values indicate better performance. Blue $\triangle$ denotes real outliers, orange $\square$ denotes synthetic outliers, and red $\star$ is our POP. `near' and `far' indicate the degree of difference between ID and OOD data.}
    \label{fig:oe_vs_pop}
\vspace{-3mm}
\end{figure}

In this paper, we introduce a novel framework, Prototypical Outlier Proxy (POP), which enables the model to learn about unknowns without exposing it to real or synthesized outliers.
As shown in Fig. \ref{fig:teaser} (right), POP, which acts as a virtual class center, can attract nearby OOD data and thereby compress the decision boundaries to mitigate the over-confidence of the deep model.
First, we transform the learnable classifier into a fixed one by using the hierarchical structure of ID data. Then, we add prototypical outlier proxies to the fixed classifier to form an OOD-aware deep model.
On the other hand, outlier proxies only cover the inter-class regions. For OOD data that are substantially different and easier to detect, 
we introduce adaptive penalties according to the severity of misclassification. 
Specifically, we propose a hierarchical similarity boundary loss (HSBL) which enables the deep model to classify data with significantly different features using the semantic hierarchical prior knowledge. 
As shown in Fig. \ref{fig:oe_vs_pop}, POP achieves balanced and excellent results in both near-OOD and far-OOD cases. 
By contrast, OE methods, such as OE~\cite{DBLP:conf/iclr/HendrycksMD19} and NPOS~\cite{DBLP:conf/iclr/TaoDZ023} perform well on near-OOD data but fail to obtain good results on the much simpler MNIST dataset. 
This is because OE methods may induce the feature extractor to overly focus on local feature discrimination between ID and OOD while lacking a global understanding of the data manifold, making it hard to perceive distant data.

In experiments, POP outperforms state-of-the-art methods, including both OE and post-hoc OOD detection, in two small-scale benchmarks and one large-scale benchmark.
We also test two of the latest challenging OOD datasets, SSB-hard \cite{ssb} and NINCO \cite{ninco}. 
Notably, POP achieves average FPR95 reductions of 7.70\%, 6.30\%, and 5.42\% over the second-best methods on CIFAR-10, CIFAR-100, and ImageNet-200, respectively. Moreover, compared to the recent method NPOS~\cite{DBLP:conf/iclr/TaoDZ023}, which relies on outlier synthesis, POP trains 7.2$\times$ faster and performs inference 19.5$\times$ times faster. 

In summary, this paper makes the following contributions:
\begin{itemize}
    \item We first identify the efficiency and generality problems of existing OE methods. To solve them, we introduce a new perspective -  Prototypical Outlier Proxy (POP) - to serve as a general surrogate for OOD data.
    \item We introduce a non-learnable classifier to mitigate the mutual influence between ID and OOD prototypes, and a similarity-based optimization objective to adaptively penalize misclassification.
    \item We conduct extensive experiments to understand the efficacy and efficiency of POP and also verify its scalability on the large-scale ImageNet dataset.
\end{itemize}

    

%% file: sections/3.motivation.tex
\section{Motivation of Algorithm Design}
\label{motivation}

We address the challenges of appending outliers during training through the use of outlier proxies. This section outlines our motivation for this approach.
Our approach is grounded in the concept of neural collapse~\cite{papyan2020prevalence}, observed during deep model training. As training progresses, features for each class converge around their mean, forming symmetrically distributed clusters. Concurrently, the classifier's weights align with these means, effectively matching well-trained features to their class prototypes. These prototypes represent the domain center of each corresponding class. Building on this, we incorporate additional prototypes as outlier proxies to create a virtual OOD domain, thereby enhancing the model's ability to recognize OOD data without being biased toward specific outlier characteristics.
However, accurately positioning outlier proxies in high-dimensional space is non-trivial. They must maintain a suitable distance from ID prototypes—neither too distant nor too close. Additionally, since ID prototypes continuously change during training, determining exact outlier proxies becomes intractable. To address this, we propose pre-defining ID prototypes by fixing the final classifier's weights, making them non-learnable. This approach simplifies the determination of suitable outlier proxies, which will be detailed in the next section.
\begin{figure}[t!] 
\vspace{-5mm}
    \centering
    \subfloat[]{\includegraphics[width=0.34\linewidth]{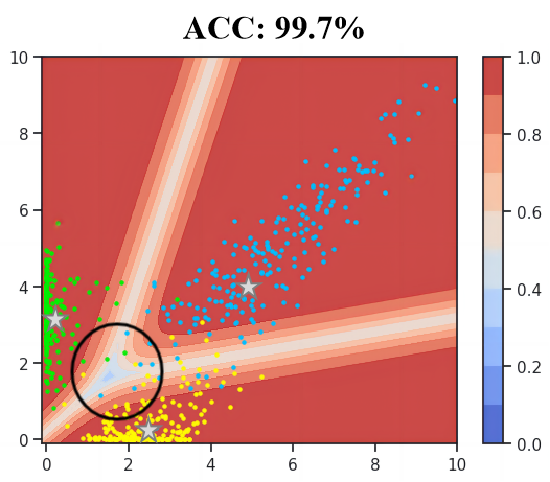} \label{sub:1}}
    \subfloat[]{\includegraphics[width=0.34\linewidth]{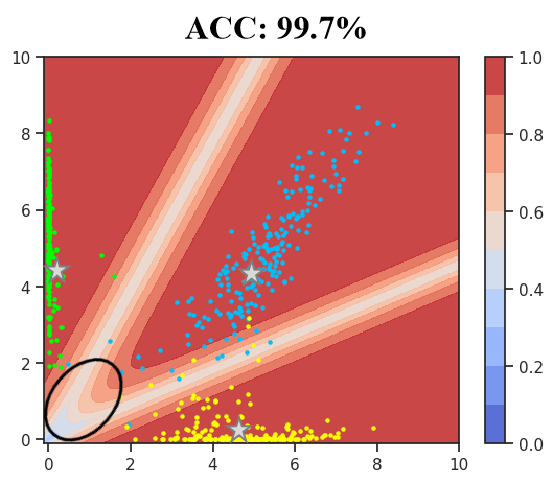} \label{sub:2}}
    \subfloat[]{\includegraphics[width=0.34\linewidth]{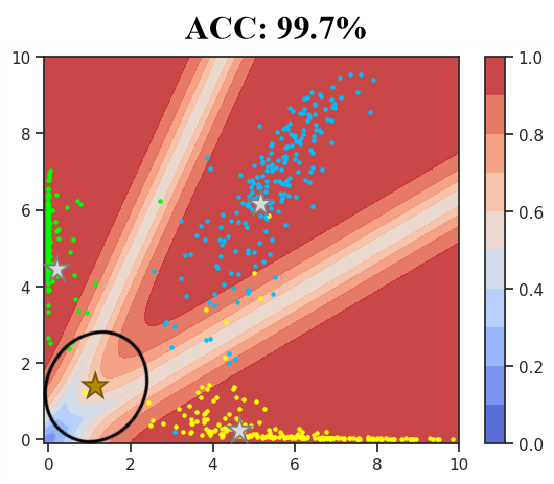} \label{sub:3}}
    \caption{\textbf{Toy example.} Use a ResNet-18 with a feature layer size of 2 for three CIFAR-10 classes. The $x$- and $y$-axes represent the feature values in the square region. We evaluate prediction confidence for each point in these classes. Yellow, green, and blue points represent deer, horse, and ship, respectively. (a) Vanilla ResNet-18. (b) Fixed ResNet-18. (c) Fixed ResNet-18 with one outlier proxy (brown star marker).}
    \label{fig:toy_example}
    \vspace{-4mm}
\end{figure}

To validate the feasibility of this intuitive idea, we conducted a toy experiment. For detailed settings of the toy experiment, please refer to Appendix A.
First, we train a baseline vanilla ResNet-18~\cite{He_2016_CVPR}. As shown in Fig~\ref{fig:toy_example} (a), this model exhibits extensive high-confidence regions (red) across the feature space, except at the decision boundaries. Even within the intersection of the three classes (black circle), the prediction confidence remains around 60\%, highlighting the prevalent issue of overconfidence in deep neural networks.
Next, we conduct an experiment using a model with pre-defined ID prototypes based on a simple semantic hierarchy before fixing the classifier of the vanilla ResNet-18. The results in Fig.~\ref{fig:toy_example} (b) show that the fixed model has tighter compression at decision boundaries, improving feature separation. However, the confidence levels in the high-confidence regions (red) and the intersection of the three classes (black circle) remain unchanged, still demonstrating overconfidence.
Finally, we blend a single prototypical outlier proxy at the center of three ID prototypes. The results, depicted in Fig. ~\ref{fig:toy_example} (c), show significant changes. The decision boundaries have widened, improving feature separation and leading to a more spread-out distribution. Notably, at the intersections (black circles), the confidence color shifts to light blue, indicating a decrease in prediction confidence to around 30\%.

%% file: sections/4.methods.tex
\section{Proposed Method}
\label{Sec:4}
\begin{figure}
    \centering
    \includegraphics[width=0.9\linewidth]{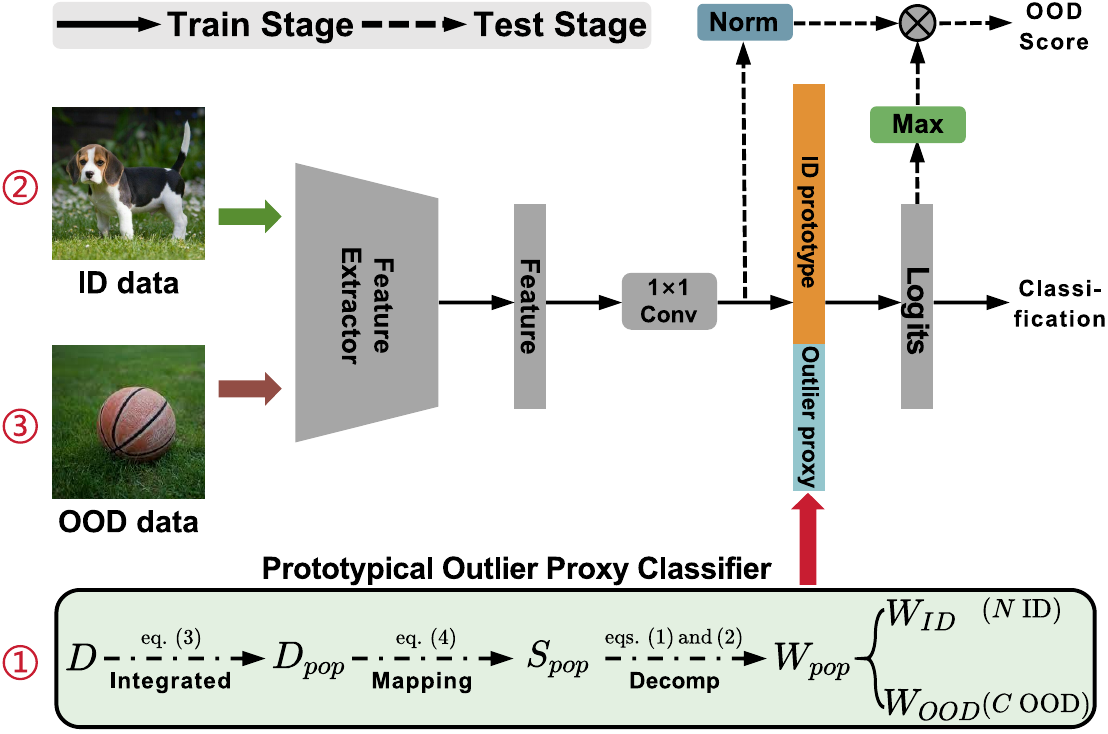}
    \caption{The overview of POP. The contributions module in POP is colored. Before training, in the green module at the bottom (\ding{172}), integrate prototypical outlier proxies into the fixed classifier. Then, ID data is fed into the model for learning (\ding{173}). Finally, during the test phase (\ding{174}), OOD data is fed into the model, and the OOD score is calculated using the feature norm and logits.}
    \label{fig:framework}
    \vspace{-4mm}
\end{figure}
In this section, we first present background knowledge on fixed classifiers, then introduce our novel approach, \textbf{POP}, a prototypical outlier proxy framework, and a hierarchical similarity boundary loss (HSBL) that imposes penalties based on misclassification severity. Finally, we explain the score function used for OOD detection.
The overview framework is illustrated in Fig.~\ref{fig:framework}.

\subsection{Preliminary: Hierarchy-Aware Frame}
\label{4.1}
HAFrame~\cite{DBLP:conf/eccv/GargSA22, DBLP:conf/iccv/LiangD23} is introduced to fix the classifier utilizing the semantic hierarchical prior of ID data. Building on the fixed ID prototypes, we can easily mix prototypical outlier proxies to determine their positions.
Common datasets and wild world data often follow a hierarchical label structure similar to WordNet~\cite{fellbaum1998wordnet}, forming a weighted tree with all class labels as leaf nodes. The semantic distance between two classes, $y_i$ and $y_j$, is measured by the height of their lowest common ancestor (LCA) in the tree, denoted as $d_{ij} = \text{H}(\text{LCA}(y_i, y_j))$, where $\text{H}(\cdot)$ is the height function, $i, j \in \{1, 2, \ldots, N\}$, and $N$ is the total number of leaf nodes. 
Next, we apply a monotonically decreasing function $\phi$ to transform $d_{ij}$ into a similarity measure. This function maps $d_{ij}$ to the interval $[0, 1]$, defining the similarity between $y_i$ and $y_j$ as $s_{ij} = \phi(d_{ij})$. Using these similarity values, we can construct a symmetric matrix $\bm{S} \in \mathbb{R}^{N\times N}$, where each element $\bm{S}_{ij} = \bm{S}_{ji} = s_{ij}$ represents the pairwise similarity between samples.
HAFrame utilizes this similarity matrix $\bm{S}$ and introduces  a set of unit vectors $\mathbf\{\bm{w_i}\}_{i=1}^N \in \mathbb{R}^N$, where each $\bm{w_i}$ has a magnitude of 1 (\underline{i.e.,} $\left || \bm{w_i} |\right |=1$). Their cosine similarity satisfies:
\begin{equation}\label{eq1}\nonumber
    \cos(\theta_{ij}) = \frac{\bm{w}_i^T \bm{w}_j}{\lVert \bm{w}_i \rVert \lVert \bm{w}_j \rVert} = \bm{w}_i^T \bm{w}_j = s_{ij}, \;
    \forall 1 \leq i \leq j \leq N.
\end{equation}
Finally, let $\bm{W}=[\bm{w_1}, \bm{w_2}, ..., \bm{w_N}]$ represent the classifier's weight vectors, which we consider as ID prototypes. The bias terms $\bm{b}$ of the linear layer were removed. Employing spectral decomposition and QR decomposition, we obtain:
\begin{equation}
\bm{S} = \bm{QPQ^T}  = \bm{(QP^\frac{1}{2}U^T) (UP^\frac{1}{2}Q^T)} = \bm{W^TW},
\label{eq1}
\end{equation}
where $\bm{Q}$ and $\bm{P}$ come from the eigenvalue decomposition of $\bm{S}$, and $\bm{U}$ is an orthogonal matrix obtained through QR decomposition from $\bm{P}$.
The ID prototypes $\bm{W}$ are given by:
\begin{equation}
    \bm{W} = \bm{UP^\frac{1}{2}Q^T}.
    \label{eq2}
\end{equation}
\subsection{Prototypical Outlier Proxy Classifier}
\label{4.2}
Using HAFrame, we obtain ID prototypes and incorporate prototypical outlier proxies into the fixed classifier. The overall process of appending outlier proxies is illustrated at the bottom
of Fig. \ref{fig:framework}. This is achieved by augmenting the ID distance matrix $\bm{D} \in \mathbb{R}^{N\times N}$ with distances greater than $d_{max} = \max(\bm{D})$, where $\bm{D}_{ij} = \bm{D}_{ji} = d_{ij}$.
To accommodate $C$ outlier proxies, we expand $\bm{D}$ to form $\bm{D}_{pop} \in \mathbb{R}^{(N+C)\times (N+C)}$ by interpolating OOD distances $d$. The structure of $\bm{D}_{pop}$ is illustrated below\footnote{For simplicity, only two OOD prototypes are shown in this example.}:
\vspace{3mm}
\begin{equation}
    \renewcommand\arraystretch{1.5}
    \bm{D}_{pop} =
    \scalebox{0.7}{ 
        $
        \begin{pNiceMatrix}
              0      & d_{12} & \cdots & d_{1N} &  d & d \\
           d_{21} & 0      & \cdots & d_{2N} &  d & d \\
          \vdots & \vdots & \ddots & \vdots &  \vdots & \vdots  \\
          d_{N1} & d_{N2} & \cdots & 0  &  d & d \\
            d   & d   & \cdots & d  &   0 & d   \\
             d    & d    & \cdots & d &  d & 0 \\
        \CodeAfter
        \OverBrace[shorten=-0.2em,yshift=0.3em]{1-1}{1-4}{\raisebox{0.25em}{$N$ ID}}
        \OverBrace[shorten=-0.2em,yshift=0.3em]{1-5}{1-6}{\raisebox{0.25em}{$C$ OOD}}
        \end{pNiceMatrix}$
    }
    \label{eq3}
\end{equation}
Then, We transform $\bm{D}_{pop}$ into a similarity matrix $\bm{S}_{pop}$ using an inverse mapping function $\phi$:
The formula for $\phi$ is:
\begin{equation}
    \phi(d_{ij}) = \frac{1}{d_{ij}+1 } ,
    \label{eq4}
\end{equation}
where $d_{ij}$ is an element of the $\bm{D}_{pop}$.
This function maps $\bm{D}_{pop}$ to the interval $[0, 1]$, resulting in $\bm{S}_{pop} = \phi(\bm{D}_{pop})$, represents the similarity between mixed ID prototypes and outlier proxies. Subsequently, utilizing the matrix decomposition from Eqs. \eqref{eq1} and \eqref{eq2}, we derive the classifier $\bm{W}_{pop}$ by combining ID prototypes $\bm{W}_{ID}$ and outlier proxies $\bm{W}_{OOD}$:
\begin{equation}
    \begin{split}
    \bm{W}_{pop} &= [\bm{W}_{ID}, \bm{W}_{OOD}] \\
    &= [\bm{w}_1, \dots, \bm{w}_N, \bm{w}_{N+1}, \bm{w}_{N+2}, \dots, \bm{w}_{N+C}].
    \end{split}
    \label{eq6}
\end{equation}
\subsection{Hierarchical Similarity Boundary Loss}
\label{4.3}
\begin{figure}[tbp!]
    \centering
    \includegraphics[width=0.67\linewidth]{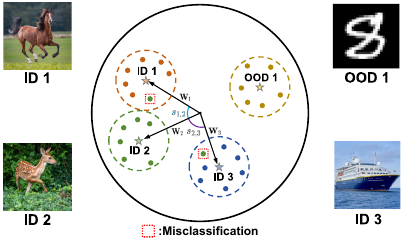}
    \caption{The principles of HSBL}
    \label{fig:loss}
    \vspace{-6mm}
\end{figure}
Due to the removal of the classifier's bias $\bm{b}$ and normalization of $\bm{W}_{pop}$ (ensuring $||\bm{w}_i||=1$), we also normalize the feature $\bm{x}$ such that $||\bm{x}||=1$, where $\bm{x} \in \mathbb{R}^M$ denotes the feature vector of the ID data. 
For the $i^{th}$ ID data's feature $\bm{x}_i$ with ground-truth label $y_i$ and predicted label $\hat{y_i}$, the cross-entropy (CE) loss $\mathcal{L}_{ce}$ in cosine space is:
\begin{equation}
\begin{split}
    \mathcal{L}_{ce} &= -\sum_i^{N+C}\log\frac{e^{{\bm{w}_{y_i}^T \bm{x}_i}}}{ {\textstyle \sum_{j=1}^{N+C}} e^{{\bm{w}_j^T\bm{x}_i}}} \\
    &= -\sum_i^{N+C}\log\frac{e^{\| \bm{w}_{y_i} \|\left \| \bm{x}_i \right \|  \cos(\theta_{y_i, i})}}{ {\textstyle \sum_{j=1}^{N+C}} e^{{\| \bm{w}_{y_i} \|\left \| \bm{x}_i \right \|\cos(\theta_{j, i})}}}\\
  &= -\sum_i^{N+C}\log\frac{e^{\cos(\theta_{y_i, i})}}{ {\textstyle \sum_{j=1}^{N+C}} e^{{\cos(\theta_{j, i})}}},
\end{split}
\label{eq6}
\end{equation}
where $\theta_{i, j}$ denotes the angle between $\bm{w}_i$ and $\bm{w}_j$. CE loss treats all misclassifications equally, however, in a hierarchical structure, misclassifying different species is more severe than misclassifying different objects within the same species (\underline{e.g.,} in autonomous driving, misclassifying a person as a sedan is far more dangerous than misclassifying a truck as a sedan, so the penalty for the former should be higher). As shown in Fig. \ref{fig:loss}, if a sample from ID 2 is misclassified as ID 1 (red dashed box), the penalty is $m_{12} = 1 - s_{12}$, which is smaller due to their high similarity $s_{12}$. Conversely, misclassifying it as ID 3 incurs a larger penalty $m_{23}$ because of the lower similarity $s_{23}$. 
We utilize the hierarchical similarity $s_{ij}$ between classes, combining it with the CE loss to improve the model's discrimination of significantly different OOD data. We integrate $s_{ij}$ into Eq. \eqref{eq6} to derive our new hierarchical similarity boundary loss:
\begin{equation}
\begin{split}
   \mathcal{L}_{hsbl}  &= -\sum_i^{N+C}\log\frac{e^{\beta({\bm{w}_{y_i}^T \bm{x}_i} - m_{\hat{y}_iy_i})}}{e^{\beta({\bm{w}_{y_i}^T \bm{x}_i} - m_{\hat{y}_iy_i})} + {\textstyle \sum_{j=1, j \ne y_i}^{N+C}} e^{\beta\bm{w}_j^T\bm{x}_i}}\\
  &=  -\sum_i^{N+C}\log\frac{e^{\beta(\cos(\theta_{y_i, i}) - m_{\hat{y}_iy_i})}}{e^{\beta(\cos(\theta_{y_i, i}) -m_{\hat{y}_iy_i})} +  {\textstyle \sum_{j=1}^{N+C}} e^{{{  \beta}\cos(\theta_{i, j})}}},
\end{split}
\label{eq7}
\end{equation}
where:
\begin{align}
m_{\hat{y_i}y_i} &= 1 - s_{\hat{y_i}y_i}
&=
\begin{cases}
0, &\quad \hat{y}_{i} = y_i\\
1 - s_{\hat{y_i}y_i} &\quad  \hat{y}_{i} \ne y_i
\end{cases}
.
\end{align}
The penalty $m_{\hat{y_i}y_i}$ is inversely proportional to the similarity between the predicted and true classes. $\beta$ is a scaling factor to enhance learning performance. 
\begin{algorithm}[tb]
\caption{The algorithm of POP}
\label{alg:algorithm}
\textit{\# The Training Stage}\\
\textbf{Input}: {Initial parameters $\theta$ for feature extractor $h(\cdot \,; \theta)$, hierarchical distance matrix $\bm{D}$, the number of outlier proxies $C$, OOD distance $d$}
\begin{algorithmic}[1] 
\STATE Insert $C^{th}$ rows and columns $d$ into $\bm{D}$ to construct $\bm{D}_{pop}$ following Eq. ~\eqref{eq3} 
\STATE Map $\bm{D}_{pop}$ through $\phi$ to get $\bm{S}_{pop}$ using Eq. ~\eqref{eq4}
\STATE Decompose $\bm{S}_{pop}$ through matrix decomposition to obtain $\bm{W}_{pop}$ following Eqs. ~\eqref{eq1} and ~\eqref{eq2}
\FOR{\textit{some training iterations}}
\STATE Optimize the parameters $\theta$ in a feature extractor $h(· ; \theta)$ using the HSBL following Eq. ~\eqref{eq7}
\ENDFOR
\RETURN $\theta, \bm{W}_{pop}$ \\
\end{algorithmic}
\vspace{2mm}
\textit{\# The Test Stage}\\
\textbf{Input}: A trained feature extractor $h(\cdot \,; \theta)$, test sample $X_i$, threshold $\lambda$
\begin{algorithmic}[1] 
\STATE Extract the feature $\bm{x}_i = h(X_i;\theta)$
\STATE Calculate OOD score $S$ using Eq. ~\eqref{eq9}
\RETURN {$\text{OOD detection decision } \bm{1}\{S \ge \lambda \} $}
\end{algorithmic}
\label{algorithm1}
\end{algorithm}
\subsection{OOD Score at Test-Time}
During the OOD detection phase, to avoid the distortion caused by softmax compression of the logits from the introduced prototypical outlier proxies, we use MaxLogit~\cite{maxlogit} instead of the softmax-based MSP~\cite{DBLP:conf/iclr/HendrycksG17}. Since the logits are in cosine space, we use the feature norm for scaling:
\begin{equation}
    S(X_i) = ||\bm{x}_i|| \cdot \max(\bm{z}_i),
\label{eq9}
\end{equation}
where $\bm{x}_i$ represents the feature vector of the $i^{th}$ sample $X_i$, and $\bm{z}_i$ denotes the logit values.
It is worth noting that the score function does not require access to ID data, making it both efficient and secure.
The training and inference stages of POP are summarized in Algorithm~\ref{algorithm1}.

%% file: sections/5.experiments.tex
\section{Experiments}
We first describe the experimental setup and then show that POP performs competitively in OOD detection compared to other state-of-the-art methods. Next, we perform extensive ablations to understand the impact of appending prototypical outlier proxies.
\begin{table*}[!ht]
    \centering
\renewcommand{\arraystretch}{1.3} 
    \setlength{\tabcolsep}{4pt}
    \resizebox{\textwidth}{!}{
     \begin{tabular}{lcccccccc}
    \toprule
     & & \multicolumn{7}{c}{\textbf{OOD Datasets}}\\
           \cmidrule(lr){3-9} 
     \multirow{1}{*}{\textbf{Methods}} & \textbf{Venue} & \textbf{CIFAR-100} & \textbf{TIN} & \textbf{MNIST} & \textbf{SVHN} & \textbf{Textures} & \textbf{Places365} & \textbf{Average}\\
     & & FPR95$\bm{/}$AUROC & FPR95$\bm{/}$AUROC & FPR95$\bm{/}$AUROC & FPR95$\bm{/}$AUROC & FPR95$\bm{/}$AUROC & FPR95$\bm{/}$AUROC & FPR95$\bm{/}$AUROC\\
    \midrule
    MSP ~\cite{DBLP:conf/iclr/HendrycksG17} &ICLR'17     &59.89$\bm{/}$86.73 & 47.21$\bm{/}$88.64 & 19.22$\bm{/}$93.95 &24.22$\bm{/}$91.57& 40.42$\bm{/}$89.13& 41.83$\bm{/}$89.35& 38.79$\bm{/}$89.90\\
    Energy ~\cite{DBLP:conf/nips/LiuWOL20} &NeurIPS'20&72.69$\bm{/}$85.55 & 62.41$\bm{/}$88.31 & 15.49$\bm{/}$96.32 & 30.16$\bm{/}$92.38&60.22$\bm{/}$88.64 & 56.37$\bm{/}$89.64 & 49.55$\bm{/}$90.14\\
    KNN ~\cite{DBLP:conf/icml/SunM0L22KNN} &ICML'22&37.90$\bm{/}$89.75&31.18$\bm{/}$91.65&20.61$\bm{/}$94.41&20.88$\bm{/}$92.89&24.50$\bm{/}$93.02&29.50$\bm{/}$92.10&27.43$\bm{/}$92.30\\
    MaxLogit ~\cite{maxlogit} & ICML'22 & 62.14$\bm{/}$86.84& 50.71$\bm{/}$88.87& 16.39$\bm{/}$95.68&31.44$\bm{/}$92.47&49.40$\bm{/}$89.38&46.21$\bm{/}$89.84 &42.71$\bm{/}$90.51\\
    ViM ~\cite{wang2022vim} &CVPR'22  &53.61$\bm{/}$87.44 & 42.49$\bm{/}$89.57 & 18.04$\bm{/}$94.25 & 18.71$\bm{/}$94.39& 21.79$\bm{/}$94.76 &44.48$\bm{/}$89.14 &33.18$\bm{/}$91.59\\
    VOS ~\cite{DBLP:conf/iclr/DuWCL22VOS} &ICLR' 22&56.21$\bm{/}$87.42 &47.18$\bm{/}$89.17&21.72$\bm{/}$94.06& 59.16$\bm{/}$83.49 &42.84$\bm{/}$89.46&44.14$\bm{/}$89.89&45.20$\bm{/}$88.92\\
    NPOS ~\cite{DBLP:conf/iclr/TaoDZ023} & ICLR'23 &37.32$\bm{/}$88.87 &30.48$\bm{/}$91.50 & 21.61$\bm{/}$94.82 & 2.54$\bm{/}$99.30 & 23.37$\bm{/}$94.34 & 30.07$\bm{/}$91.86 & \underline{24.23}$\bm{/}$\underline{93.44}\\
    \rowcolor{gray!30}
    POP (Ours) & N/A & 32.19$\bm{/}$91.77 & 21.18$\bm{/}$94.76 &2.96$\bm{/}$99.43&7.72$\bm{/}$98.45&16.59$\bm{/}$96.50  &18.56$\bm{/}$95.65&\textbf{16.53}$\bm{/}$\textbf{96.09}\\
    \bottomrule
    \end{tabular}
    }
    \caption{Experiment results on CIFAR-10. The utilized metrics include FPR95 (\(\downarrow\)), aiming for lower values to indicate better performance; AUROC (\(\uparrow\)), where higher values denote superior discriminative ability; and ID Accuracy, measuring the rate of correct classifications. The top-performing models are marked with \textbf{bold} for the best and \underline{underline} for the second best.}
    \label{cifar10}
\end{table*}

\subsection{Experimental Setup}
\textbf{Datasets.} 
For comprehensive experiments, we adopt the OpenOOD
\footnote{https://github.com/Jingkang50/OpenOOD} 
benchmark ~\cite{yang2022openood, zhang2023openood}, which provides an accurate, standardized, and unified evaluation for fair testing. We include small-scale datasets CIFAR-10~\cite{cifar} and CIFAR-100~\cite{cifar}, and the large-scale ImageNet-200, which is a subset of ImageNet-1k ~\cite{ImageNet} with the first 200 classes, as our ID datasets.
Among them,
(\romannumeral1) CIFAR-10 is a small dataset with 10 classes, including 50k training images and 10k test images. We establish OOD test dataset with \underline{CIFAR-100}, \underline{Tiny ImageNet} (TIN) ~\cite{tin}, \underline{MNIST} ~\cite{mnist2} (including \underline{Fashion MNIST} ~\cite{mnist1}), \underline{Texture}\cite{textures}, and \underline{Places365} ~\cite{places365}.
(\romannumeral2) CIFAR-100, another small dataset, consists of 50k training images and 10k test images, with 100 classes. The OOD test dataset includes \underline{CIFAR-10}, with the remaining datasets configured identically to those in (\romannumeral1).
(iii) For the large-scale dataset ImageNet-200, the OOD test dataset consist of \underline{SSB} ~\cite{ssb} \underline{NINCO} ~\cite{ninco}, \underline{iNatruelist} ~\cite{van2018inaturalist}, \underline{Place365}, and \underline{OpenImage-O} ~\cite{wang2022vim}.
\\
\textbf{Baselines.} 
We compare our POP with 7 baselines. They are mainly divided into two categories: (1) post-hoc inference methods: \textbf{MSP} ~\cite{DBLP:conf/iclr/HendrycksG17}, \textbf{Energy} ~\cite{DBLP:conf/nips/LiuWOL20}, \textbf{ViM} ~\cite{wang2022vim}, and \textbf{Maxlogit} ~\cite{maxlogit}; (2) adding outliers methods:  \textbf{VOS} ~\cite{DBLP:conf/iclr/DuWCL22VOS},
\textbf{NPOS} ~\cite{DBLP:conf/iclr/TaoDZ023}.
\\
\textbf{Evaluation metrics.}
We evaluate our method using (1) the false positive rate (FPR95) at the threshold where the true positive rate for ID samples is 95\% and (2) the area under the receiver operating characteristic curve (AUROC). Both metrics are reported as percentages. In ablation experiments, FPR95 and AUROC are averaged across the benchmark.
\\
\textbf{Training details.}
We train a ResNet-18 model ~\cite{He_2016_CVPR} from scratch for 100 epochs on CIFAR-10 and CIFAR-100, and 90 epochs on ImageNet-200, using a single Nvidia 4090. Training is performed with the SGD optimizer, a learning rate of 0.1, momentum of 0.9, and weight decay of 0.0005. 
The complete experimental setup is provided in Appendix B.
\vspace{1mm}
\subsection{Main Results}
\label{sec:exp;subsec:main}

\begin{table*}[t]
    \centering
\renewcommand{\arraystretch}{1.3} 
    \setlength{\tabcolsep}{4pt}
    \resizebox{\textwidth}{!}{
     \begin{tabular}{lcccccccc}
    \toprule
     & & \multicolumn{7}{c}{\textbf{OOD Datasets}}\\
           \cmidrule(lr){3-9} 
     \multirow{1}{*}{\textbf{Methods}} & \textbf{Venue} & \textbf{CIFAR-10} & \textbf{TIN} & \textbf{MNIST} & \textbf{SVHN} & \textbf{Textures} & \textbf{Places365} & \textbf{Average}\\
     & & FPR95$\bm{/}$AUROC & FPR95$\bm{/}$AUROC & FPR95$\bm{/}$AUROC & FPR95$\bm{/}$AUROC & FPR95$\bm{/}$AUROC & FPR95$\bm{/}$AUROC & FPR95$\bm{/}$AUROC\\
    \midrule
    MSP ~\cite{DBLP:conf/iclr/HendrycksG17} &ICLR'17     &59.10$\bm{/}$78.54&50.36$\bm{/}$82.30&63.47$\bm{/}$73.54&56.08$\bm{/}$79.10& 61.37$\bm{/}$78.06& 55.41$\bm{/}$79.62& 57.63$\bm{/}$78.52\\
    Energy ~\cite{DBLP:conf/nips/LiuWOL20} &NeurIPS'20&58.82$\bm{/}$79.01& 52.14$\bm{/}$82.66 &57.57$\bm{/}$77.30& 51.24$\bm{/}$82.40&60.27$\bm{/}$79.33& 56.54$\bm{/}$79.82 & 56.09$\bm{/}$80.08\\
    KNN ~\cite{DBLP:conf/icml/SunM0L22KNN} &ICML'22&72.41$\bm{/}$76.76&49.64$\bm{/}$83.18&44.21$\bm{/}$83.67&56.14$\bm{/}$82.65&51.92$\bm{/}$83.86&61.46$\bm{/}$78.79&55.96$\bm{/}$\underline{81.48}\\
    MaxLogit ~\cite{maxlogit} & ICML'22 & 58.42$\bm{/}$79.33 & 50.89$\bm{/}$82.97 & 48.64$\bm{/}$80.40 & 51.97$\bm{/}$83.23 & 61.62$\bm{/}$78.52 & 58.60$\bm{/}$79.49 & 55.02$\bm{/}$80.65\\
    ViM ~\cite{wang2022vim} &CVPR'22  &71.17$\bm{/}$71.73&54.71$\bm{/}$77.94&46.87$\bm{/}$81.76& 44.52$\bm{/}$84.25& 46.99$\bm{/}$86.28&60.49$\bm{/}$76.17 &\underline{54.12}$\bm{/}$79.68\\
    VOS ~\cite{DBLP:conf/iclr/DuWCL22VOS} &ICLR' 22&59.79$\bm{/}$78.69 &54.29$\bm{/}$82.00&43.98$\bm{/}$84.34 & 75.66$\bm{/}$73.30 &66.58$\bm{/}$76.66&58.37$\bm{/}$79.29&59.77$\bm{/}$79.04\\
    NPOS ~\cite{DBLP:conf/iclr/TaoDZ023} &ICLR' 23&70.97$\bm{/}$75.72 &54.17$\bm{/}$81.60&77.73$\bm{/}$70.96& 31.40$\bm{/}$91.72 &50.90$\bm{/}$84.19&59.80$\bm{/}$78.53& 57.49$\bm{/}$80.45\\
    \rowcolor{gray!30}
    POP (Ours) & N/A &66.92$\bm{/}$76.74&51.74$\bm{/}$82.46 & 31.38$\bm{/}$91.29 &30.91$\bm{/}$89.76&53.20$\bm{/}$83.11&52.80$\bm{/}$80.64 &\textbf{47.82}$\bm{/}$\textbf{84.00}\\
    \bottomrule
    \end{tabular}
    }
    \caption{Experiment results on CIFAR-100.}
    \label{cifar100}
\end{table*}
In the following, we present the performance of POP.  \\
\textbf{Results on CIFAR-10.} On the CIFAR-10 dataset, the results in Tab. ~\ref{cifar10} show that POP outperforms other methods, leading by a significant margin in most OOD datasets. Particularly, on the challenging CIFAR-10, TIN, and Place365 datasets, where features closely resemble those in the ID dataset, POP achieves over 91\% AUROC. POP also performs well on structured OOD datasets like MNIST, SVHN, and Textures, indicating that our method is effective across OOD data of varying difficulty.
In contrast, other methods, whether post-hoc or mixing in outliers, fail to achieve an AUROC above 90\% on any OOD dataset.
Our average OOD AUROC performance surpasses the second-best by \textbf{2.05\%}, while FPR95 is significantly reduced by \textbf{5.4\%}. It is worth noting that POP not only excels in AUROC but also shows exceptional performance in FPR95, as observed in CIFAR-100 and ImageNet-200. This highlights POP's strong robustness across different evaluation metrics and datasets.
\begin{table*}[t]
    \centering
\renewcommand{\arraystretch}{1.2} 
    \setlength{\tabcolsep}{4pt}
    \resizebox{\textwidth}{!}{
     \begin{tabular}{lccccccc}
    \toprule
     & \multicolumn{7}{c}{\textbf{OOD Datasets}}\\
           \cmidrule(lr){3-8} 
     \multirow{1}{*}{\textbf{Methods}} & \textbf{Venue} & \textbf{SSB-hard} & \textbf{NINCO} & \textbf{iNaturelist} & \textbf{Places365} & \textbf{OpenImage-O} & \textbf{Average}\\
     & & FPR95$\bm{/}$AUROC & FPR95$\bm{/}$AUROC & FPR95$\bm{/}$AUROC & FPR95$\bm{/}$AUROC & FPR95$\bm{/}$AUROC & FPR95$\bm{/}$AUROC\\
    \midrule
    MSP ~\cite{DBLP:conf/iclr/HendrycksG17} &ICLR'17     &65.76$\bm{/}$79.92&43.59$\bm{/}$85.91&26.87$\bm{/}$92.67&41.61$\bm{/}$88.51&35.80$\bm{/}$88.89&\underline{42.73}$\bm{/}$87.17\\
    Energy ~\cite{DBLP:conf/nips/LiuWOL20} &NeurIPS'20 &69.44$\bm{/}$79.32&49.59$\bm{/}$85.04&26.83$\bm{/}$92.51 &35.86$\bm{/}$90.14&38.04$\bm{/}$88.90&43.95$\bm{/}$\underline{87.18}\\
    KNN ~\cite{DBLP:conf/icml/SunM0L22KNN} &ICML'22& 72.48$\bm{/}$77.24&48.41$\bm{/}$85.36& 28.44$\bm{/}$92.57 &45.56$\bm{/}$85.47&37.30$\bm{/}$88.66&46.44$\bm{/}$85.85\\
    MaxLogit ~\cite{maxlogit} & ICML'22 &70.81$\bm{/}$79.95& 51.31$\bm{/}$85.34 & 25.90$\bm{/}$92.85 & 35.90$\bm{/}$90.30 & 36.50$\bm{/}$89.42 & 44.49$\bm{/}$87.72\\
    ViM ~\cite{wang2022vim} &CVPR'22   &69.78$\bm{/}$75.49&45.81$\bm{/}$82.91 &30.00$\bm{/}$88.96 &39.99$\bm{/}$85.10&36.66$\bm{/}$86.70 &44.45$\bm{/}$83.83\\
    VOS ~\cite{DBLP:conf/iclr/DuWCL22VOS} &ICLR'22 &70.86$\bm{/}$78.91&52.00$\bm{/}$84.21&26.96$\bm{/}$92.82 & 51.57$\bm{/}$82.93 & 38.01$\bm{/}$88.98 &47.88$\bm{/}$85.57\\
    NPOS ~\cite{DBLP:conf/iclr/TaoDZ023} &ICLR'23& 73.61$\bm{/}$74.19& 48.53$\bm{/}$84.67&20.67$\bm{/}$94.75&46.99$\bm{/}$88.08& 29.39$\bm{/}$91.57& 43.84$\bm{/}$86.65\\
    \rowcolor{gray!30}
    POP (Ours) & N/A   & 66.71$\bm{/}$78.09 &43.48$\bm{/}$86.82&15.84$\bm{/}$96.09&29.24$\bm{/}$91.78&31.30$\bm{/}$90.72&\textbf{37.31}$\bm{/}$\textbf{88.70}\\
    \bottomrule
    \end{tabular}
    }
    \caption{Experiment results on ImageNet-200.}
    \label{imagenet200}
\end{table*}
\\
\textbf{Results on CIFAR-100.} On the CIFAR-100 dataset, as shown in Tab. ~\ref{cifar100}. Compared to the second-best result, POP performs exceptionally well, improving by \textbf{6.3\%} in FPR95 and \textbf{3.52\%} in AUROC. POP exceeds 80\% in AUROC across all datasets, except CIFAR-10. This is because CIFAR-100 is a more fine-grained dataset, and many of its labels overlap with those in CIFAR-10 due to the hierarchical structure shared between the two datasets. The performance of VOS and NPOS, which introduced feature-based synthetic outliers, is poor on the simple MNIST dataset. This suggests that using prototypical outlier proxies, rather than actual outliers, is more flexible and effective for handling OOD detection.
\\
\textbf{Results on ImageNet-200.}
On the large-scale ImageNet-200 dataset, as detailed in Tab. \ref{imagenet200}, POP maintains excellent generalization performance, achieving competitive results on the challenging SSB ~\cite{ssb} and NINCO ~\cite{ninco} datasets. SSB only includes semantic shift, and NINCO ensures that none of the objects in its dataset have appeared in ImageNet (ID), but their features are very similar to ID. VOS and NPOS perform poorly on NINCO, even worse than post-hoc methods that require no training. This suggests that the unreliability of adding outliers is limited and may only be effective on certain OOD datasets.
In contrast, compared to the second-best method, POP reduces FPR95 by \textbf{5.32\%}, highlighting its exceptional ability to minimize false positives and enhance detection reliability. The average AUROC is also improved by \textbf{1.52\%}, underscoring POP's strong and consistent performance across diverse OOD scenarios.
\subsection{Ablation Study}
\begin{table}[t]
\centering
\scalebox{0.9}{%
\belowrulesep=0pt
\aboverulesep=0pt
\begin{tabular}{l|c c c|c c}
\toprule
 ID dataset   &F &O &H& FPR95 $\downarrow$ & AUROC $\uparrow$ \\
 \hline 
\multirow{4}{*}{{CIFAR-10}}
 & \cmark & & & 27.45 & 92.75\\
 & \cmark & & \cmark& 25.65 & 92.72\\
 &\cmark &\cmark & & 23.90 & 93.67  \\
 &\cmark &\cmark &\cmark & \textbf{21.25} & \textbf{94.69} \\

\hline 
\multirow{4}{*}{{CIFAR-100}}  
 &\cmark &  &  &54.56 & 79.94  \\
 & \cmark & & \cmark& 53.28 & 81.10\\
 &\cmark &\cmark & &  52.10 &81.62  \\
 &\cmark &\cmark &\cmark & \textbf{47.82}  & \textbf{84.00}  \\
\bottomrule
\end{tabular}%
}
\caption{The ablation study results for CIFAR-10 and CIFAR-100. The best performances in \textbf{bold}. F: fixed ResNet-18. O: fixed ResNet-18 with outlier proxies. H: update using HSBL.}
\label{tab:ablation2}
\vspace{-5.5mm}
\end{table}
To better understand POP, we conducted a thorough ablation study, detailed in Tab. \ref{tab:ablation2}. (1) (F + H) Using a fixed model with HSBL for parameter updates notably improves performance on CIFAR-100, but shows no significant improvement on CIFAR-10. We argue that the more complex hierarchical structure of CIFAR-100 provides more similarity information between classes. (2) (F + O) Appending prototypical outlier proxies enhances generalization performance, showing improvements on both CIFAR-10 and CIFAR-100 compared to using only the fixed model. (3) (F + O + H) Appending prototypical outlier proxies with HSBL for parameter updates, there is a substantial improvement compared to previous methods. 
Outlier proxies build virtual OOD domains, preserving the model's semantic space and alleviating over-confidence in deep models. Meanwhile, HSBL helps the model classify samples by discriminative features, enhancing its ability to recognize distant OOD data. Thus, their combination improves the model's generalization.
\subsection{Effects of Hierarchy}
We compared prototypes that were randomly orthogonalized with those decomposed based on hierarchical distances, as shown in Tab. \ref{tab:random}. The results indicate that using random prototypes, which lack hierarchical prior information, yields sub-optimal performance. 
\begin{table}[!tbp]
    \centering
    \scalebox{0.95}{
    \begin{tabular}{lcc}
        \toprule
         \multirow{2}{*}{Fix Method} & CIFAR-10 & CIFAR-100 \\
         & FPR95$\bm{/}$AUROC & FPR95$\bm{/}$AUROC \\
        \midrule
        Random &32.43$\bm{/}$90.48& 63.06$\bm{/}$74.44\\
        Hierarchy &\textbf{16.53$\bm{/}$96.09}&\textbf{47.82$\bm{/}$84.00}\\ 
        \bottomrule
    \end{tabular}
    }
    \caption{The results of various fixed methods on CIFAR-10 and CIFAR-100.}
    \label{tab:random}
\end{table}
\vspace{-2mm}
\subsection{Analysis of the HSBL}
\begin{figure*}[!h]
	\centering
	\small
	\setlength\tabcolsep{1mm}
	\renewcommand\arraystretch{0.1}
	\begin{tabular}{cccc}
		\includegraphics[scale=1]{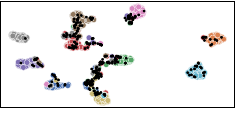} &
		\includegraphics[scale=1]{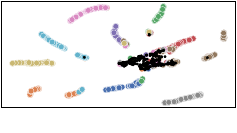} & 
		\includegraphics[scale=1]{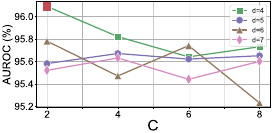} &
		\includegraphics[scale=1]{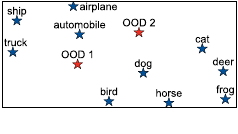} \\
		(a) CE loss & (b) HSBL & (c) The analysis of $d$ and $C$ & (d) Prototypes\\
	\end{tabular}
        \vspace{1mm}
	\caption{The analysis of HSBL loss and POP. ID test data (CIFAR-10) and OOD data (SVHN) features are visualized using UMAP ~\cite{UMAP}, with ResNet-18 trained with CE loss and HSBL. Colored points represent ID data, and black points represent OOD data. The prototypes, the classifier's weight vectors, are visualized using UMAP.}
	\label{fig:umap}
    \vspace{-2mm}
\end{figure*} 
The high-dimensional feature learned by CE loss and HSBL is visualized as shown in Figs. \ref{fig:umap} (a) and (b). Clearly, using HSBL results in tighter intra-class compression and greater separation on inter-class. HSBL is capable of compressing the features of OOD data into a smaller region. This helps the model better distinguish between ID and OOD data. Also, recent work ~\cite{ma2022principles, NEURIPS2023_29f35148} indicates that models with better compression enhance generalization. 
\subsection{Impact of Prototypical Outlier Proxies}
On CIFAR-10, with the maximum ID hierarchical distance \( d_{max} = 3 \), we set $d \in \{4, 5, 6, 7\}$ and $C \in \{2, 4, 6, 8\}$ in Eq. \eqref{eq3} and conducted a grid search. The results are shown in Fig. \ref{fig:umap} (c).
As the number of outlier proxies increases, the overall performance shows a downward trend. We speculate that may be because, in a simple hierarchical structure like CIFAR-10, the inclusion of an excessive number of outlier proxies could prevent the model from effectively utilizing the hierarchical prior information.
Therefore, we visualized all the prototypes for the best-performing combination, which is $d=4$ and $C=2$, as shown in Fig. \ref{fig:umap} (d). It can be observed that the ID prototypes (blue star markers) in the upper-left maintain their semantic structure and are categorized as \textit{tools}. In the lower-right, \textit{animal} categories are represented, with deer and horse being relatively close. The two outlier proxies (red star markers) successfully separate the two categories.
This structured separation helps the model better capture the intrinsic semantic information of the data.
\vspace{-2mm}
\subsection{Analysis of Time Efficiency for Testing}
Tab. \ref{tab:mean_std} shows the comparison of running times between POP and other methods for integrating outliers during the training and inference phases. During training, POP is 2.40 times faster than VOS and 7.23 times faster than NPOS. In testing, POP achieves 1.06 times the speed of VOS and 19.50 times the speed of NPOS.
This efficiency is due to POP's simple modifications to the vanilla model and because post-hoc methods avoid accessing ID data. In contrast, NPOS is slower due to the use of KNN \cite{DBLP:conf/icml/SunM0L22KNN} for distance computation on ID data. The efficiency and effectiveness of POP pave the way for applications requiring high real-time performance.
\begin{table}[!tbp]
    \centering
    \scalebox{0.95}{%
    \belowrulesep=0pt
    \aboverulesep=0pt
    \begin{tabular}{lccc}
        \toprule
        Metric (s) & VOS & NPOS & POP (Ours) \\
        \midrule
        Train time &$21.58 \pm 0.42 $& $65.03 \pm 4.82$& $\textbf{9.00} \pm \textbf{1.12}$ \\
       Infer time &$5.84 \pm 0.39$ & $107.05 \pm 3.48$ & $\textbf{5.49} \pm \textbf{0.32}$\\
        \bottomrule
    \end{tabular}
    }
    \caption{Comparison of running times: training on CIFAR-10 and testing on Place365, conducted on an NVIDIA RTX 4090 (each method tested over 5 rounds on the full dataset).}
    \label{tab:mean_std}
    \vspace{-3mm}
\end{table}

%% file: sections/6.related_work.tex
\section{Related Work}
\subsection{OOD Detection Methods}
In OOD detection, one category of methods involves using post-processing techniques without training. Techniques like MSP ~\cite{DBLP:conf/iclr/HendrycksG17} focus on the classifier's output probabilities. Methods such as MaxLogit ~\cite{maxlogit} and energy scores ~\cite{DBLP:conf/nips/LiuWOL20} use the logits. The Mahalanobis distance measure ~\cite{lee2018simple}, relies on the classifier's feature representations. Another approach involves retraining a model by incorporating outliers. OE ~\cite{DBLP:conf/iclr/HendrycksG17} directly trains with labeled real outliers. UDG ~\cite{DBLP:conf/iccv/YangWFYZZ021} and MCD ~\cite{Yu2019UnsupervisedOD} use unlabeled real outliers for unsupervised training. Dream-OOD ~\cite{dream} employs a powerful diffusion model to generate synthetic outliers based on a text-conditioned latent space derived from ID data. VOS ~\cite{DBLP:conf/iclr/DuWCL22VOS} assumes a certain distribution in the feature space to generate outliers, while NPOS~\cite{DBLP:conf/iclr/TaoDZ023} extends VOS by generating outliers without a specific distribution. MODE~\cite{DBLP:journals/tip/ZhangGHHSS23} proposes a multi-scale framework that combines global and local features to improve out-of-distribution detection performance. VOso~\cite{nie2024out} proposes a novel approach to address DNN overconfidence in out-of-distribution detection by creating virtual outliers through semantic region perturbation of in-distribution samples. 
In contrast to these methods, our POP approach directly addresses OOD detection from the perspective of outlier proxies. 
\subsection{Fixed Classifier}
Early research into optimizing memory and computational resources has explored fixing the classifier. \cite{DBLP:conf/iclr/HardtM17} examines modifying the final layer of deep learning models, while \cite{DBLP:conf/iclr/HofferHS18} suggests fixing the classifier to a global scale constant and demonstrates the feasibility of starting the classifier from a Hadamard matrix. 
FRCR ~\cite{FRCR} first fixes the classifier using a random matrix and then reorders the classifier for continual learning. \cite{tao} proposed a simplex equiangular tight frame has achieved promising results in fixing classifiers on long-tailed datasets.
Meanwhile, HAFrame ~\cite{DBLP:conf/iccv/LiangD23} employs a hierarchical structure to fix the classifier, improving performance on fine-grained classification tasks.
Building on HAFrame, we leverage hierarchical prior information of ID data and introduce outlier proxies to address the OOD detection task.

%% file: sections/7.discussion.tex
\section{Conclusion}
In this paper, we introduce a simple yet effective OOD detection framework, POP, which reshapes the decision boundary between ID and OOD data without exposing the model to real or synthetic OOD data.
By doing so, POP prevents the model from being influenced by specific outliers or learning biased characteristics. Moreover, it eliminates the need for synthetic samples, greatly improving both training and inference speeds.
Extensive experiments on multiple benchmark datasets demonstrate the superiority of the proposed POP.

\newpage

\section{Acknowledgments}

This work was supported in parts by NSFC (U21B2023), ICFCRT (W2441020), Guangdong Basic and Applied Basic Research Foundation (2023B1515120026), and Scientific Development Funds from Shenzhen University.

%% file: sections/appendix.tex
\cleardoublepage
\onecolumn  
\appendix   
\section{Appendix} 
\vspace{5mm}
\subsection{A \quad Toy Experiment Training Details}
\label{toy}
The toy experiment (a) in Fig. ~\ref{fig:toy_example},
We trained a standard ResNet18 for 30 epochs on three classes and modified the size of the feature layer to 2 for visualization purposes. We utilized a cross-entropy loss and SGD optimizer with a momentum of 0.9, setting a learning rate of 0.1 and weight decay of 0.0005.

Regarding the toy experiment (b), we set the distance from deer to horse as $1$, and the distance from deer and horse to ship as $3$. The distance matrix $\bm{D}$ for these three classes is as follows:
\begin{equation*}
\renewcommand\arraystretch{1.5}
    \bm{D} = \begin{Bmatrix}
  0& 1 &3\\
  1& 0 &3 \\
  3& 3 &0
\end{Bmatrix}.
\end{equation*}

Next, we transform $\bm{D}$ into hierarchical similarity \(\bm{S}\) using Eq. (4) and finally obtain the classifier weight vector \(\bm{W}\) to fix the classifier by decomposing $\bm{S}$ using Eqs. (1) and (2). The training settings are identical to the toy experiment (a).

In toy experiment (c), we constructed the prototypical outlier proxy distance matrix  $\bm{D}_{pop}$ by adding an outlier proxy with a distance of $d=4$ to the distance matrix $\bm{D}$ of these three classes. Its formula is as follows,
\begin{equation*}
\renewcommand\arraystretch{1.5}
    \bm{D}_{pop} = \begin{Bmatrix}
  0& 1 & 3 &4\\
  1&  0&3 & 4\\
  3& 3 &0 & 4\\
  4& 4 & 4 & 0
\end{Bmatrix}.
\end{equation*}
Finally, the contaminated classifier was obtained by decomposing $\bm{D}_{pop}$, with training settings identical to those used previously.

\subsection{B\quad Experiment Details}
\subsubsection{B.1 \quad Dataset}
The specific semantic label tree of the CIFAR-10 dataset is shown in Fig. 7. The distance between the ``\textit{Cat}" and ``\textit{Horse}" is the distance to their lowest common ancestor (LCA), ``\textit{Animal}", which is 2.
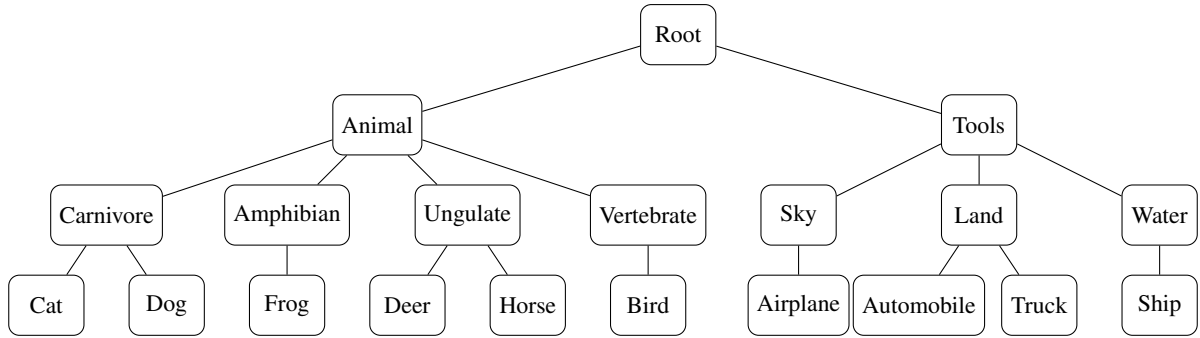
\begin{figure}[htbp]
\begin{center}
\begin{tikzpicture}[
    scale=0.8,
    level 1/.style={sibling distance=100mm},
    level 2/.style={sibling distance=30mm},
    level 3/.style={sibling distance=20mm},
    every node/.style={draw, rounded corners, minimum width=10mm, minimum height=8mm, text centered, align=center, font=\small}
]

\node {Root}
    child { node {Animal}
        child { node {Carnivore}
            child { node {Cat} }
            child { node {Dog} }
        }
        child { node {Amphibian}
            child { node {Frog} }
        }
        child { node {Ungulate}
            child { node {Deer} }
            child { node {Horse} }
        }
        child { node {Vertebrate}
            child { node {Bird} }
        }
    }
    child { node {Tools}
        child { node {Sky}
            child { node {Airplane} }
        }
        child { node {Land}
            child { node {Automobile} }
            child { node {Truck} }
        }
        child { node {Water}
            child { node {Ship} }
        }
    };

\end{tikzpicture}
\end{center}
\caption{The hierarchical structure of CIFAR-10.}
\label{fig:hierarchy-tree}
\end{figure}

The maximum hierarchical distances $d_{max}$ for CIFAR-10, CIFAR-100, and ImageNet-200 are presented in Tab. 7. POP calculates the OOD distance based on $d_{max}$, thereby introducing outlier prototypes.
\begin{table}[htbp]
\centering
\begin{tabular}{@{}cccc@{}}
\toprule
 & CIFAR-10 & CIFAR-100 & ImageNet-200 \\
\midrule
$d_{max}$ &3 &5 &15  \\
\bottomrule
\end{tabular}
\caption{
The maximum hierarchical distance across different datasets.}
\label{tab:two-by-four}
\end{table}
\newpage
\subsubsection{B.2 \quad Training Details}
In this section, we present the implementation details and experimental results for our method trained from scratch. Our evaluation covers three datasets: CIFAR-10, CIFAR-100, and ImageNet-200. We outline the training configurations of our method in Tab. \ref{tab:train}.
\begin{table}[ht!]
  \centering
  \small
  \begin{tabular}{lccc}
    \toprule
     & CIFAR-10 & CIFAR-100 & ImageNet-200 \\
    \midrule
    Training epochs & 100   & 100   & 90 \\
    Momentum & 0.9   & 0.9   & 0.9 \\
    Batch size & 128   & 128   & 128 \\
    Weight decay & 0.0001 & 0.0001 & 0.0001 \\
    Initial LR & 0.1   & 0.1   & 0.1 \\
    LR schedule & cosine & cosine & cosine \\
    Scaling factor of HSBL $\beta$ & 10   & 5   & 10 \\
    The number of prototypical outlier proxy $C$  & 2   & 60   & 40 \\
    OOD distance $d$ & 4   & 7   & 18 \\
    \bottomrule
    \end{tabular}%
  \caption{Configurations of POP.}
  \label{tab:train}%
\end{table}%

\subsubsection{B.2 \quad Effect of $\beta$} We validated the scaling factor $\beta$ in Eq. (7), and the results are shown in Tab. 9. On CIFAR-10 and CIFAR-100, it can be observed that the performance is mediocre when $\beta$ is small, but it decreases when $\beta$ becomes too large.
This experiment demonstrates that selecting an appropriate scaling factor for datasets of different scales helps the model converge faster, learn better features, and improve generalization.
\begin{table}[ht]
    \centering
    \begin{tabular}{lcc}
        \toprule
        \multirow{2}{*}{$\beta$} & CIFAR-10 & CIFAR-100 \\
         & FPR95$\bm{/}$AUROC & FPR95$\bm{/}$AUROC \\
        \midrule
         1&23.72$\bm{/}$94.19& 54.65$\bm{/}$80.53\\
        5 &18.92$\bm{/}$95.40&\textbf{47.82$\bm{/}$84.00}\\ 
        10 &\textbf{16.53$\bm{/}$96.09}&53.12$\bm{/}$81.93\\ 
        20 &18.69$\bm{/}$95.45&57.47$\bm{/}$79.91\\ 
        \bottomrule
    \end{tabular}
    \caption{The results of different $\beta$ values on CIFAR-10 and CIFAR-100.}
    \label{tab:random}
\end{table}